\documentclass{article}
\usepackage{spconf,amsmath,graphicx}

\title{Small-footprint Keyword Spotting Using Deep Neural Network and Connectionist Temporal Classifier}
\name{Zhiming Wang, Xiaolong Li, Jun Zhou\sthanks{\{zhiming.wzm, xl.li, jun.zhoujun\}@antfin.com}}
\address{AI Department, Ant Financial Group, Hangzhou, China}

\begin{document}
%
\maketitle
\begin{abstract}
Mainly for the sake of solving the lack of keyword-specific data, we propose one Keyword Spotting (KWS) system using Deep Neural Network (DNN) and Connectionist Temporal Classifier (CTC) on power-constrained small-footprint mobile devices, taking full advantage of general corpus from continuous speech recognition which is of great amount. DNN is to directly predict the posterior of phoneme units of any personally customized key-phrase, and CTC to produce a confidence score of the given phoneme sequence as responsive decision-making mechanism. The CTC-KWS has competitive performance in comparison with purely DNN based keyword specific KWS, but not increasing any computational complexity.
\end{abstract}
\begin{keywords}
Keyword Spotting, Deep Neural Network, Connectionist Temporal Classifier, Embedded Speech Recognition
\end{keywords}

\section{Introduction}
\label{sec:format}

As mobile devices are widely used, speech-related technologies are becoming more commonplace. For example, the increasingly popular conversational assistants, such as Apple's Siri, Microsoft's Cortana and Amazon's Alexa, all utilize speech interaction to enable desired user experience.

Among them, one important voice interactive technology is called Keyword Spotting (KWS), which continuously listens for specific keywords in an audio stream to initiate voice input. The KWS system, which incorporates the interface for specific query commands, is especially useful in situations like driving, and has extensive product applications, like ``Okey Google'' in the smart speaker of Google Home. To prevent the devices disconnecting from the remote server for spotting tasks, the KWS system should run on portable devices, and therefore must satisfy the requirements of high accuracy, low latency, small memory footprint and low computing cost.

With great successes of Deep Neural Network (DNN) in large vocabulary continuous speech recognition (LVCSR)~\cite{hinton2012deep},
the DNN based Deep-KWS system~\cite{chen2014small}
has shown to outperform the traditional keyword and filler Hidden Markov Models (HMMs) in the literatures~\cite{rose1990hidden, wilpon1991improvements},
which can be computationally expensive depending on the HMM's topology. The Deep-KWS is trained to predict sub-word or full-word units in key phrases, and then gives a confidence score based on frame-level label posteriors in DNN's output layer. Then the work~\cite{sainath2015convolutional}
substitutes Convolutional Neural Network (CNN) for DNN in small-footprint KWS tasks by means of limiting the number of multiplications or parameters. These methods heavily rely on keyword-specific speech data, resulting in a limited number of training samples and a big man-powered or economic cost for data acquisition. For lack of a mass of training data, the model of the larger number of parameters would have worse performance. These shortcomings impose restrictions on flexibility in practical deployment.

If not having enough keyword-specific data or even without them, we reflect how to set up one efficient KWS system appropriate for commercial products. Could the available LVCSR corpus play a role in this aspect? The answer is yes. In this paper, we propose one keyword spotting system using Deep Neural Network and Connectionist Temporal Classifier (CTC)~\cite{graves2006connectionist},
which makes full use of general LVCSR corpus. We refer it as CTC-KWS. One DNN is trained to directly predict the posterior of connotative phoneme units of the keywords, followed by a CTC which gives a confidence score of the phoneme sequence as responsive decision-making mechanism. Both the CTC-KWS and Deep-KWS have almost the same number of parameters, and hence equal computational complexity. The CTC-KWS differs from previous proposed Recurrent Neural Network (RNN) based keyword spotting~\cite{fernandez2007application}
in that the latter employed a limited number of key-phrases as modeling targets, and bidirectional Long Short Term Memory (LSTM) with more computing cost.  The CTC-KWS adopts flexible decision-making mechanism, breaks the bottle-neck of being short of keyword-specific data, and supports any personally customized trigger words, which is more convenient in practice. We observe that CTC is good at sequence labeling tasks, which explains why the CTC-KWS produces competitive performance in comparison with the baseline Deep-KWS. Using adaptive training method on some but still limited quantities of key-phrase specific data, we could achieve better performance result in the CTC-KWS.

 \begin{figure*}[t]
    \centering
    \includegraphics[scale = 0.1]{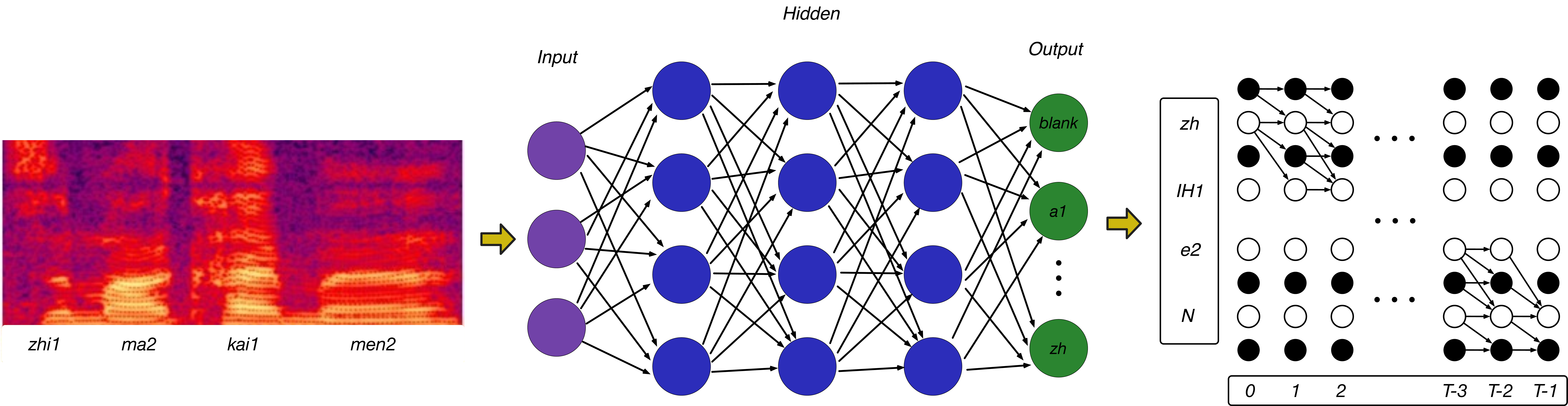}
    \caption{The proposed keyword spotting system, components from left to right are: (i) Feature Extraction, (ii) Deep Neural Network, and (iii) Connectionist Temporal Classifier where black circle indicates blank unit.}
    \label{fig: the proposed KWS system}
 \end{figure*}

The rest of this paper is organized as follows: in Section 2, we describe our proposed keyword spotting system, including how to train and deploy the CTC-KWS model, then the experimental details follow in Section 3, and Section 4 closes with the conclusion and outlook.

\section{Proposed Keyword Spotting System}
\label{sec:format}

Our proposed keyword spotting system is illustrated in Fig.~\ref{fig: the proposed KWS system}. The system framework consists of three major components: (i) a feature extraction module to process coming speech signal, (ii) a Deep Neural Network to encode acoustic features into more abstract representations, and (iii) on top of the output layer of DNN, a Connectionist Temporal Classifier to produce scores for the implicit phoneme sequence of given keyword. In what follows, we describe these in detail, inclusive of model training and deployment.

\subsection{Feature Extraction and Modeling Units}
\label{ssec:subhead}

The feature extraction is identical to that in LVCSR. We generate 40-dimensional log filter bank energies as acoustic features over a window of 25 milliseconds overlapped every 10 milliseconds. Contiguous frames of features are stacked together to allow for more discriminative information from both left and right context, which are asymmetric for reducing latency, that is, 5 future frames and 10 frames in the past in our experiment. Apply cepstral mean and variance normalization on each dimension of stacked features. To save computing cost, only run our proposed algorithm in voice regions perceived by one voice activity detection system.

All tone phonemes are taken into account as modeling units. Context independent or dependent phonemes are befitting, the latter are of bigger sizes and more generalized to unseen train data, but to reduce computation in the last layer of DNN, we just consider CI phonemes in our experiment where there are 72 Chinese CI phoneme units including one blank unit for CTC. Once keywords are personally customized by users, their corresponding phoneme sequence is determined by lexicon, which is the target label sequence of CTC scoring.

\subsection{Deep Neural Network}
\label{ssec:subhead}

Each frame of acoustic feature vector is fed into a feed-forward fully connected DNN. DNN establishes a function mapping relationship between acoustic features ${\bf x}^{0} \in \Re^{n_{0}}$ and modeling units ${j}$ in the last layer as follows:
 \begin{eqnarray}
     {\bf x}^{i} &=& \sigma({\bf x}^{i-1} {\bf \widehat W}^{i} + {\bf B}^{i}), 1 \leq i \leq N, \label{eq1} \\
     {\bf z} &=& {\bf x}^{N} {\bf \widehat W}^{N + 1} + {\bf B}^{N + 1}, \label{eq2} \\
     y_{j} &=& \frac{\exp({\bf z}_{j})}{\sum_k \exp({\bf z}_{k})}. \label{eq3}
 \end{eqnarray}
Here, the matrix or vector is bold, the subscript indexes associated element, the widehat signifies matrix transpose. ${\bf x}^{i, i > 0} \in \Re^{n_{i}}$ is the output of ${ith}$ hidden layer,  ${\bf W}^{i} \in \Re^{n_{i} \times n_{i - 1}}$, ${\bf B}^{i} \in \Re^{n_{i}}$ are the weight and bias parameters of ${ith}$ layer, $n_{i}$ is the number of nodes of ${ith}$ layer, set ${\theta = \{{\bf W}^{i}, {\bf B}^{i} | 1 \leq i \leq N + 1}\}$, ${N}$ is the total number of hidden layers. $\sigma$ is the element-wise nonlinear operator such as the function of rectified linear unit (ReLU) $\sigma(a) = \max(a, 0)$~\cite{zeiler2013rectified}. The equation (\ref{eq3}) is a softmax, representative of the estimated posterior of label unit ${j}$.

Recent years witness many successful end-to-end applications of LSTM based RNN with CTC in LVCSR~\cite{graves2013speech, miao2015eesen}.
However, our experiment shows that under resource constraints, CTC could also be combined with feed-forward DNN, not limited to RNN. DNN is used here to meet minimum compute and footprint requirement of mobile device, and hence the number of DNN's parameters is modest for reduced complexity, approximately hundreds of nodes in each hidden layer.

\subsection{Connectionist Temporal Classifier}
\label{ssec:subhead}

As its name suggests, CTC was specifically designed for sequence labeling tasks~\cite{graves2006connectionist}.
Unlike the cross entropy criterion where the frame-level alignments between input signal features and label sequences are known, the CTC objective is to automatically learn the alignments. This removes the need for pre-segmenting data (for example, forced alignment), and the lengths of inputs and labels are not necessarily equal. Given the modeling label units are drawn from ${L}$, CTC is on top of the softmax layer of DNN which consists of ${|L|}$ units and an additional blank unit. The introduction of the blank unit relieves the burden of making label predictions since there is no symbol being emitted at a frame when it is uncertain. The output activations of DNN are used to define a probability distribution over all possible lattice paths of length up to that of the input sequence.

Define $y_{j}^{t} (j \in [0, |L|], t \in [0, T))$ as the probability that DNN outputs element ${j}$ at time step ${t}$, given the input sequence $x^{T}$ of frame length ${T}$ and the label sequence ${l^{\leq T}}$, $l_{i} \in L$. A CTC path $\pi = (\pi_{0}, \cdot\cdot\cdot, \pi_{T - 1})$ is a sequence of tokens at the frame level. It differs from ${l}$ in that the CTC path allows repetitions of non-blank labels and occurrences of the blank unit. By removing first the repeated labels and then the blanks, the CTC path ${\pi}$ can be mapped to its corresponding label sequence ${l}$. For example, $\tau($``$aa-b-c$''$) = \tau($``$abb-cc-$''$) = $ ``$abc$'', where the many-to-one mapping function is defined as ${\tau}$, and ``$-$'' indicates the blank. Assumed the output probabilities at each time step to be conditionally independent, the probability of the path ${\pi}$ is
 \begin{equation}
    p(\pi|x; \theta) = \prod_{t = 0}^{T - 1} y_{\pi_{t}}^{t}.
    \label{eq4}
 \end{equation}
 Then the likelihood of ${l}$ can be calculated by summing the probabilities of all the paths mapped onto it by ${\tau}$,
 \begin{equation}
    p(l|x; \theta) = \sideset {} {_{\pi \in \tau^{-1}(l)}} \sum p(\pi|x; \theta).
    \label{eq5}
 \end{equation}
 However, summing over all the CTC paths is computationally intractable. To solve this, A. Graves~\cite{graves2006connectionist}
 proposed a forward-backward dynamic programming algorithm, where all the possible CTC paths are compactly represented as a trellis, like the rightmost illustration of Fig.~\ref{fig: the proposed KWS system}.

 When training, the CTC objective is
 \begin{equation}
    \theta^{*} = argmin_{\theta} \sideset {} {_{(x, l) \in S}} \sum -log(p(l|x; \theta))
    \label{eq6}
 \end{equation}
 using the training data set ${S}$. Coincidentally when decoding, in a short time span (for example, a window in size of 100 frames and overlapped every 25 frames), the spotting engine will make a positive decision once the confidence score, given as $log(p(l|x; \theta^{*}))$, is greater than some predefined threshold, which is tuned on one development data set. Just one tuned hyper-parameter brings great convenience.

\subsection{Model Training and Deployment}
\label{ssec:subhead}

In practice, the DNN and CTC model is trained on GPU servers using Nesterov's acceleration~\cite{DBLP:conf/icml/SutskeverMDH13}
asynchronous stochastic gradient descent method. The network parameters are randomly initialized with a uniform distribution on the range of (-0.02, 0.02), the initial learning rate is 0.008, the momentum is 0.9. Use cross validation on one development data set to half decay the learning rate or determine whether the training is convergent. Currently, some open source deep learning toolkits fully support DNN and CTC model training, for instance, Google's Tensorflow\footnote{https://www.tensorflow.org}, or alternatively Kaldi\footnote{http://kaldi-asr.org} based EESEN\footnote{https://github.com/yajiemiao/eesen} where the CTC part could also be further optimized using Warp-CTC\footnote{https://github.com/baidu-research/warp-ctc}.

To achieve better performance and robust model, the trick of keyword-specific adaptive training is worthy of exploring, that is, further fine tune the aforementioned universal model with smaller learning rate on a certain amount of speech data for specific keywords.

When the keyword spotting system is deployed on power-constrained and small-footprint mobile devices, the engine has a stream mechanism to process coming speech signals batch by batch in size of maximum 32 frames. Using ReLU as nonlinearity speeds up calculation~\cite{lei2013accurate},
about 50\% of neurons activated and forward propagated in each hidden layer. The additive and multiplicative operations could be further accelerated using architecture-dependent vector instructions, e.g., NEON in ARM.

\section{Experimental Details}
\label{sec:format}

In this section, we evaluate the performance of our proposed CTC-KWS system, in comparison with the keyword-specific Deep-KWS~\cite{chen2014small} as baseline.

\subsection{Data}

Considering deployed scenarios, our experiment is conducted with Mandarin Chinese corpus, but our technique is easily extended, not limited to Mandarin. For training the DNN and CTC model, we use the general Mandarin LVCSR corpus, which consists of 2.5K hours of transcribed utterances, about 1.9M ones.
 \begin{figure*}[htb]
    \begin{minipage}[htb]{0.5\textwidth}
        \raggedleft
        \includegraphics[scale = 0.5]{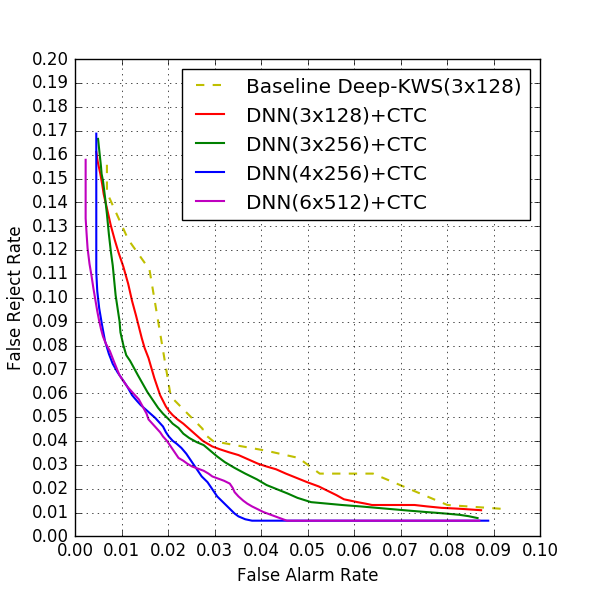}
        \caption{The performance of the Deep-KWS vs. CTC-KWS.}
        \label{fig: the_performance_of_two}
    \end{minipage}
    \begin{minipage}[htb]{0.5\textwidth}
        \raggedright
        \includegraphics[scale = 0.5]{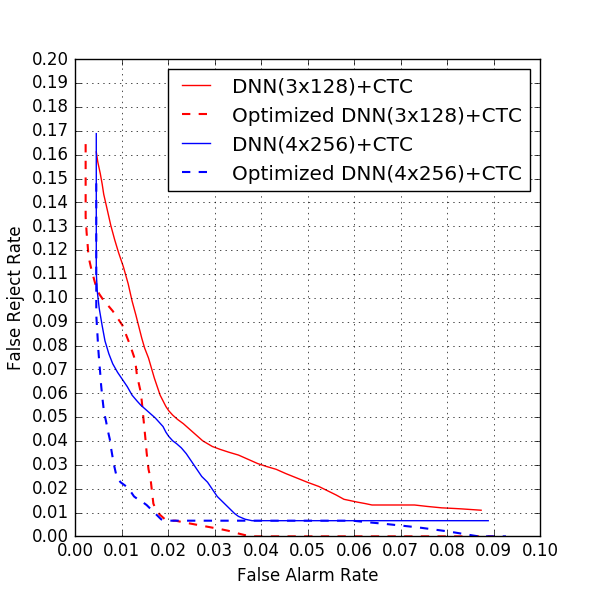}
        \caption{The optimized performance of adaptive training.}
        \label{fig: the_opt_performance}
    \end{minipage}
 \end{figure*}

In the Deep-KWS system, one keyword-specific corpus is used at training time, which includes roughly 2K examples for each keyword. In ~\cite{chen2014small}, the number of training examples per keyword is around 2.3K, and 40K for ``Ok Google''. Since keyword data is manually acquired for specific spotting tasks, especially at the start point, it is of a limited number and a waste of manpower or economic cost. Quoad hoc, the CTC-KWS is superior over the Deep-KWS. The negative examples are comprised of other short phrases excluding keywords. Table~\ref{tab:number_of_KW_samples} summarize the numbers of samples for training, development and evaluation, which are non-overlapping.

Without loss of generality, we choose 10 key phrases in Chinese: ``a1 tao2 ni3 hao3'', ``an1 na4 ni3 hao3'', ``da3 kai1 zhi1 fu4 bao3'', ``da3 kai1 shou3 tao2'', ``da3 kai1 gao1 de2'', ``da3 kai1 tao2 bao3 shou3 ye4'', ``zhi1 ma2 kai1 men2'', ``zhi1 ma2 xin4 yong4'', ``zhang4 dan1 cha2 xun2'' and ``yu2 e2 cha2 xun2''\footnote{Respectively, they mean ``Hi, Ali TaoBao'', ``Hi, AnNa, who is a chatbot in Alipay'', ``Open Alipay'', ``Open TaoBao'', ``Open navigation assistant GaoDe Map'', ``Go to www.taobao.com'', ``Good luck'', `` Credit score'', ``Check the bill in one's account'' and ``Check the balance in one's account'' in Chinese. Among them, ``Alipay'', ``TaoBao'' and ``GaoDe Map'' are well-known APPs on mobile devices in China.}. They span about 1 to 2 seconds in time.
 \begin{table}[th]
   \caption{The numbers of samples in the experiment}
   \label{tab:number_of_KW_samples}
   \centering
   \begin{tabular}{c c c}
     \hline
     \textbf{Type}\;\;\;\;\;\;\;\;  &  \textbf{\#Positive per Keyword}  &  \textbf{\#Negative} \\
     \hline
     Training\;\;\;\;\;\;\;	        &      2K	                        &        20K  \\
     Development	                &    0.8K	                        &        10K  \\
     Evaluation\;\;\;	            &    0.8K	                        &        10K  \\
     \hline
   \end{tabular}
 \end{table}

\subsection{Metrical Performance Evaluation}

The experimental results are demonstrated in the form of a modified version of Receiver Operating Characteristic (ROC) curves, where the false reject rate(that is, a key phrase is present but a negative decision is given) is on the vertical axis and the false alarm rate (that is, a key-phrase is not present but a positive decision is made) is on X-axis. Lower curves are better in performance, depending on the tradeoff between the two indices. The ROC is obtained by sweeping through confidence thresholds and averaging the curves vertically over all keywords in test.

The baseline Deep-KWS employs the settings as follows: (i) the front-end signal processing as is described in Section $2.1$ except for 30 left frames and 10 right frames in the context, (ii) one DNN with 3 hidden layers and 128 nodes per hidden layer, (iii) ReLU as the non-linearity, all of which gain better performance in ~\cite{chen2014small}. Figure~\ref{fig: the_performance_of_two} shows the performance of the Deep-KWS versus CTC-KWS of different parameter numbers. It is clear that the proposed CTC-KWS outperforms the baseline, even when it is of the least number of parameters ($N = 3, n_{i, 1 \leq i \leq N} = 128)$. Although the number of phoneme units in the CTC-KWS is greater than that of word units in the Deep-KWS, the size of input context window is opposite, therefore both systems are of nearly the same number of parameters, and hence equivalent computational complexity. When increasing DNN's depths or nodes per hidden layer, the CTC-KWS has superior performance to the baseline. Compared at false alarm rate 1.5\%, as the number of parameters becomes bigger, the performance of the CTC-KWS tends to stability. However, as is reported in~\cite{chen2014small}, the Deep-KWS of the bigger number of parameters performs worse due to lack of enough keyword-related data for model training.

\subsection{The optimized performance of adaptive training}

Figure~\ref{fig: the_opt_performance} presents the optimized performance of adaptive training when using some but still limited amount of data for specific key-phrase, as is described in Section $2.4$. Again, we observe a consistent improvement with respect to their corresponding universal model. The gains are larger at very low false alarm rate such as 1.5\%, which is a desirable operating point for practical deployment. There is no doubt that more specific data would further improve performance regardless of the Deep-KWS or CTC-KWS. But this is beyond our original intention in this paper since we focus on the deficiency of keyword-specific data.

\subsection{The automatical alignment learned by CTC}

\begin{figure}[!htb]
    \centering
    \includegraphics[width = \linewidth]{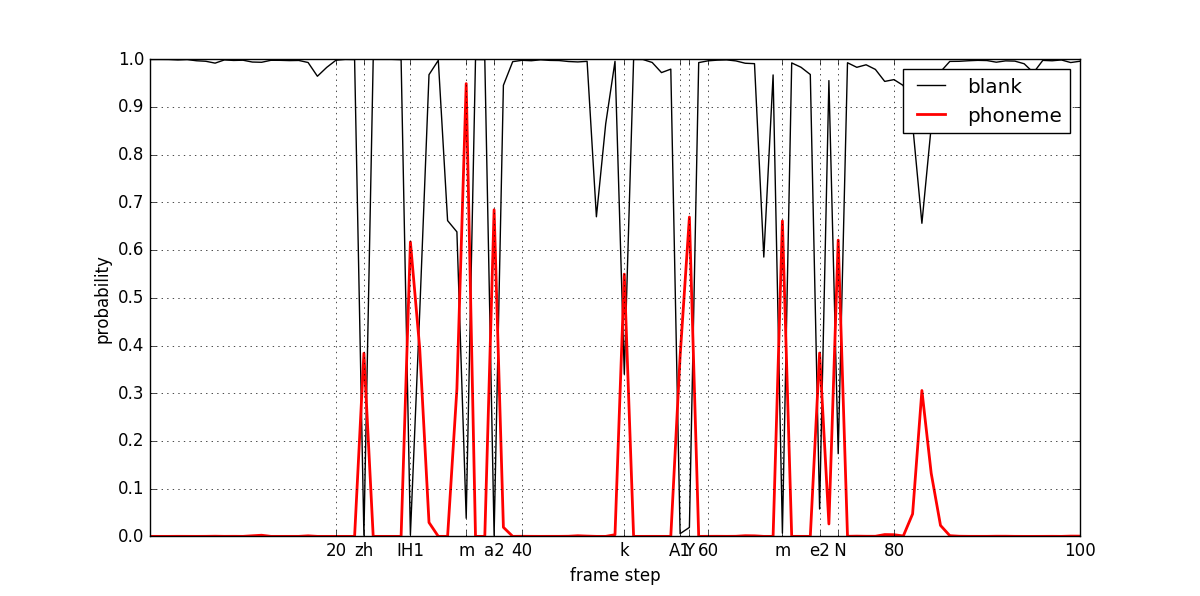}
    \caption{The automatical alignment learned by CTC.}
    \label{fig: the_alignment_learned_by_CTC}
 \end{figure}

As is illustrated in Figure~\ref{fig: the_alignment_learned_by_CTC}, CTC automatically learns the alignment between acoustic features and phonemes or the blank, where on the vertical axis are the output activations of DNN corresponding to the probabilities of observing phonemes or the blank at particular time step. The predicted phonemes take shape of a series of spikes, separated by the blank. There are blank predictions in uncertain acoustic regions. This automatical alignment shows the merit of CTC which is good at sequence labeling.

\subsection{Real Time Performance on mobile devices}

 When the CTC-KWS ($N = 4, n_{i, 1 \leq i \leq N} = 256)$ is deployed on mobile devices, The Real Time Factor (RTF) is listed in Table~\ref{tab:rtf}. In ARM A8, use NEON to optimize additive and multiplicative operations. In addition, the model's memory footprint of different complexities varies from 0.5M to 1.5M to meet the diversified demands, the smaller memory footprint, the less computing cost at runtime.
 \begin{table}[th]
   \caption{RTF of the CTC-KWS($4\times256$) on mobile devices}
   \label{tab:rtf}
   \centering
   \begin{tabular}{c c c}
     \hline
     \textbf{Architecture}  &  \textbf{Memory and Frequency}  &  \textbf{RTF} \\
     \hline
     ARM A8	                &    512M, 1G Hz	              &      0.2218         \\
     MIPS\;\;\;\;           &    128M, 1G Hz	              &      0.3\;\;\;\;\;  \\
     \hline
   \end{tabular}
 \end{table}

\section{Conclusion and Outlook}

In this paper, we consider how to utilize the available LVCSR corpus to set up the CTC-KWS system on embedded mobile devices, mainly to solve the lack of keyword-specific data. For supporting any personally customized trigger words, the CTC-KWS is more convenient and more flexible in practical deployment. We observe a larger improvement with respect to the baseline Deep-KWS, but not increasing any computing cost. Using adaptive training method on some keyword-specific data, better performance is achieved for a real need.

Here we use Deep Neural Network in our solution just for meeting minimum requirements of embedded mobile devices, however, slightly relaxing restrictions on this, especially souping up computing, the CTC-KWS is scalable to Recurrent Neural Network or Convolutional Neural Network. That is, DNN, RNN and CNN are consistent in encoding low level of signal features into higher level of distributed representations, and then according to the label posteriors in the softmax layer, CTC gives a confidence score as decision-making mechanism. We will continue to do more research on how to combine CTC with compressed RNN or CNN (like ~\cite{sainath2015convolutional}) in our proposed CTC-KWS framework in the future. In addition, the robustness of the CTC-KWS in the condition of noise or under far-field environment is also worthy of further study.

\bibliographystyle{IEEEbib}
\bibliography{strings,refs}

\end{document}